\documentclass[preprint,epsfig,onecolumn,floats, showkeys]{revtex4}
\usepackage{makeidx}
\usepackage{amsmath}
\usepackage{amssymb}
\usepackage{graphicx}
\usepackage{epsfig}

\begin{document}

\title{Apply Ant Colony Algorithm to Search All Extreme Points of Function}
\author{Chao-Yang Pang}
\email{cypang@live.com}
\email{cypang@sicnu.edu.cn}
\affiliation{Group of Gene Computation, College of Mathematics and Software Science,
Sichuan Normal University, Chengdu 610066, China\\
Group of Gene Computation, Key Lab. of Visual Computation and Virtual
Reality, Sichuan Normal University,Chengdu 610068, China; }
\author{Hui Liu}
\affiliation{Group of Gene Computation,College of Mathematics and Software Science,
Sichuan Normal University, Chengdu 610066, China}
\author{Xia Li}
\affiliation{College of Information Engineering, Shenzhen University, Shenzhen, Guangdong
province 518060, China }
\author{Yun-Fei Wang}
\affiliation{College of Computer Science, Sichuan Normal University , Chengdu 610066,
China}
\author{Ben-Qiong Hu}
\affiliation{College of Information Management, Chengdu University of Technology, 610059,
China}

\begin{abstract}
To find all extreme points of multimodal functions is called extremum
problem, which is a well known difficult issue in optimization fields.
Applying ant colony optimization (ACO) to solve this problem is rarely
reported. The method of applying ACO to solve extremum problem is explored
in this paper. Experiment shows that the solution error of the method
presented in this paper is less than $10^{-8}$.
\end{abstract}

\maketitle


\section{Introduction}

\subsection{Extremum Problem}

Multimodal function refers to the function which has more than one extreme
point. To find all extreme points is called extremum problem, which is a
well known difficult issue in optimization fields. Many practical
engineering problems can be converted as this problem, such as the detection
of multiple objects in military field. Therefore, solving the extremum
problem is a useful study topic. To solve the extremum problem, many methods
of optimization are applied, such as genetic algorithm (GA) \cite{MinQiang
Li}, simulated annealing (SA) \cite{Qing-yun Tao}, particle swarm
optimization algorithm (PSO) \cite{Li Li,Jiang Wu}, immune algorithm ( IA)
\cite{Rui-ying Zhou}, and so on. However, currently there is rare report
that applying Ant Colony Optimization (ACO) to solve the extremum problem.
The motivation of this paper is to apply ACO to search all extreme points of
function.

\subsection{Introduction of Ant Colony Optimization (ACO)}

Ant Colony Optimization (ACO) was first proposed by Dorigo (1991) \cite{M.
Dorigo1,A. Colorni,M. Dorigo2}. The inspiring source of ACO is the foraging
behavior of real ants. When ants search for food, they initially explore the
area surrounding their nest in a random manner. As soon as an ant finds a
food source, it remembers the route passed by and carries some food back to
the nest. During the return trip, the ant deposits pheromone on the ground.
The deposited pheromone, guides other ants to the food source. And the
feature has been shown, indirect communication among ants via pheromone
trails enables them to find the shortest routes between their nest and food
sources. ACO imitates this feature and it becomes an effective algorithm for
the optimization problems \cite{M. Dorigo3}. It has been successfully
applied to many combinatorial optimization problems \cite{M.O. Ball1,M.O.
Ball2,K. Doerner}, such as Traveling Salesman Problem (TSP) \cite{M.
Dorigo4,M. Manfrin}, Quadratic Assignment Problem(QAP) \cite{L. M.
Gambardella1}, Job-shop Scheduling Problem(JSP) \cite{L. M. Gambardella2},
Vehicle Routing Problem(VRP) \cite{B. Bullnheimer,P. Forsyth}, Data
Mining(DM) \cite{Rafael S} and so on.

The application of ACO pushes the study of ACO theory, and its two
main study topics are the analysis of convergence and runtime. M.
Birattari proves the
invariance of ACO and introduced three new ACO algorithms \cite{M. Birattari}%
. Convergence is one of focus study of ACO. Walter J. Gutjahr studied the
convergence of ACO firstly in 2000 \cite{W. J. Gutijahr1}. T. St$\overset{..}%
{u}$ezle and M. Dorigo proved the existence of the ACO convergence
under two conditions, one is to only update the pheromone of the
shortest route generated at each iteration step, the other is that
the pheromone on all routes has lower bound \cite{Stuezle}. C.-Y.
Pang and et.al. found a potential new view point to study ACO
convergence under general condition,
that is entropy convergence in 2009 \cite{PangACOEntropy}. In ref.\cite%
{PangACOEntropy}, the following conclusion is get: ACO may not converges to
the optimal solution in practice, but its entropy is convergent. The other
study focus of ACO is time complexity. ACO has runtime $O(tmN^{2}$), where $%
t $, $m$ and $N$ refers to the number of iteration steps, ants, cities and $%
M=[\frac{N}{1.5}]$ in general. To reduce runtime, cutting down parameter $t$
and $N$ is the main way possibly. In 2008, Walter J. Gutjahr presented some
theoretical results about ACO runtime \cite{W. J. Gutijahr2, W. J. Gutijahr3}%
. Since runtime is proportional to the square of $N$, parameter $N$
is the key factor of runtime. Through cutting down $N$ to reduce
runtime is a choice. And Pang and et al. do the following attempt
\cite{PangACOSLC}. Firstly, cluster all cities into some classes
(group) and let ACO act on these small classes respectively to get
some local TSP routes. And then joint these local route to form the
whole TSP route. If class is compact, the length of local route got
at every iteration step will change continually possibly, where
compactness refers to all data cluster in a small region tightly.
And this property result in the conclusion: the convergence
criterion $\frac{|L_{t}-L_{t+1}|}{L_{t}}\rightarrow 0$ is the
marker of ACO convergence, where $L_{t}$ is the length of local route at $%
t-th$ iteration step. Thus, minimum iteration number $t$ can be estimated by
the marker approximately.

The study of ACO theory speeds up its application again. ACO not
only can be applied to solve discrete optimization problems, but
also to continuous ones. The first method for continuous-space
optimization problems, called Continuous ACO (CACO), was proposed by
Bilchev and Parmee (1995)\cite{G A Bilchev}, and later it was used
by some others \cite{M. R. Jalali,Wodrich M,Mathur M,Jalali MR}. In
general, the application of ACO to continuous optimization problems
need to transform a continuous search space to a discrete one. Other
methods include that, in 2002, Continuous Interacting Ant Colony
(CIAC) was proposed by Dreo and Siarry \cite{Dreo J}, and in 2003,
an adaptive ant colony system algorithm for continuous-space
optimization problems was proposed by Li Yan-jun \cite{Li Y J}, and
so on.

\subsection{Framework of ACO}

Traveling Salesman Problem (TSP) is a famous combinatorial problem,
it can be stated very simply:

A salesman visit $N$ cities cyclically provided that he visits each
city just once. In what order should he visit them to minimis the
distance traveled?

The typical application of ACO is to solve TSP, and its basic idea
is stated as below:

When an ant passes through an edge, it releases pheromone on this edge. The
shorter the edge is, the bigger the amount of pheromone is. And the
pheromone induces other ants to passes through this edge. At last, all ants
select a unique route, and this route is shortest possibly.

The framework of ACO is introduced as below:

\textbf{Step1 (Initialization)}: Posit $M$ ants at different $M$ cities
randomly; Pre-assign maximum iteration number $t_{max}$; Let $t=0$, where $t$
denotes the $t-th$ iteration step; Initialize amount of pheromone of every
edge.

\textbf{Step2} While($t<t_{max}$)

\{

\ \textbf{Step2.1:} Every ant selects its next city according to the
transition probability. The transition probability from the $i-th$ city to
the $j-th$ city for the $k-th$ ant is defined as Eq.\ref{ACO probability}.

\begin{equation}
p_{ij}^{(k)}(t)=\left\{
\begin{tabular}{ccc}
$\frac{\tau _{ij}^{\alpha }(t).\eta _{ij}^{\beta }}{\underset{s\in
allowed_{k}}{\sum }\tau _{is}^{\alpha }(t).\eta _{is}^{\beta }},$ & $\ \ \
if $ & $j\in allowed_{k}$ \\
$0,$ & \multicolumn{2}{c}{$otherwise$}%
\end{tabular}%
\right.  \label{ACO probability}
\end{equation}%
, where $allowed_{k}$ denotes the set of cities that can be accessed by the $%
k-th$ ant; $\tau _{ij}(t)$ is the pheromone value of the edge $(i,j)$; $\eta
_{ij}$ is local heuristic function defined as

\begin{equation*}
\eta _{ij}=\frac{1}{d_{ij}}
\end{equation*}%
, where $d_{ij}$\ is the distance between the $i-th$\ city and the $j-th$\
city; the parameters $\alpha $ and $\beta $ determine the relative influence
of the trail strength and the heuristic information respectively.

\ \textbf{Step2.2:} After all\textbf{\ }ants finish their travels, all
pheromone values are updated according to Eq.\ref{ACO update}.

\begin{equation}
\tau _{ij}(t+1)=(1-\rho )\cdot \tau _{ij}(t)+\Delta \tau _{ij}(t)
\label{ACO update}
\end{equation}%
\begin{equation*}
\Delta \tau _{ij}(t)=\overset{m}{\underset{k=1}{\sum }}\Delta \tau
_{ij}^{(k)}(t)
\end{equation*}%
\bigskip
\begin{equation*}
\Delta \tau _{ij}^{(k)}(t)=\left\{
\begin{tabular}{ccc}
$\frac{Q}{L^{(k)}(t)},$ & $\ \ \ if$ & $the$ $k-th$ $ant$ $pass$ $edge$ $%
(i,j)$ \\
$0,$ & \multicolumn{2}{c}{$otherwise$}%
\end{tabular}%
\right.
\end{equation*}%
, where $L^{(k)}(t)$\ is the length of the route passed by the $k-th$\ ant
during the $t-th$\ iteration; $\rho $ is the persistence percentage of the
trail (thus, $1-\rho $\ corresponds to the evaporation); $Q$ denotes
constant quantity of pheromone.

\ \textbf{Step2.3:} Increase iteration step: $t\leftarrow t+1$

\}

\textbf{Step3:} End procedure and select the shortest route as output from
the routes traveled by the ants.

\section{Apply ACO to Search All Extreme Points of Function}

\subsection{Basic Idea}

Assume that the function is $f(x)$, $x$ is real number and it belongs to a
closed interval $[a,b]$. The task of this paper is to extract all extreme
points that the corresponding value is minimal locally. The basic idea of
this paper is stated roughly as below:

Divide interval $[a,b]$ into many tiny intervals with equal size. Suppose
these small interval are $\{I_{1},I_{2},\cdots ,I_{n}\}$ and the center of
interval $I_{i}$ is denoted by $x_{i}$. Suppose the neighbor interval of
interval $I_{i}$\ is $I_{i+1}$ (or $I_{i-1}$). And an ant is put at the
center of each small interval. If $f(x_{i})>f(x_{i+1})$, the ant at interval
$I_{i}$ will move to interval $I_{i+1}$\ possibly, just liking there is an
virtual edge between $I_{i}$ and $I_{i+1}$. And assume the virtual edge is $%
e(I_{i},I_{i+1})$. The weight (virtual distance) of edge $e(I_{i},I_{i+1})$\
is proportional to $f(x_{i})-f(x_{i+1})$. That is, the bigger $%
f(x_{i})-f(x_{i+1})$ is, the more possibly the ant moves to $I_{i+1}$ from $%
I_{i}$. When the ant moves to $I_{i+1}$, it releases pheromone at $I_{i+1}$,
and the pheromone is proportional to value $f(x_{i})-f(x_{i+1})$. The
pheromone depositing on $I_{i+1}$ will attract other ants move to it.

After some iteration steps, some intervals contain many ants while other
ones contain no ant. The intervals containing ants include extreme points
possibly, while other ones include no extreme point possibly. And then keep
the intervals which contain ants, and divide them into much more small
intervals, repeat the same procedure again until the size of intervals is
sufficient small. At last all ants will stay around extreme points. The
centers of these sufficient small intervals are the approximations of
extreme points.

From above discussion, it can be seen that the realization of the basic idea
consists of four parts: partition of interval $[a,b]$ and initialization,
rule of ant moving, rule of pheromone updating, and keeping the intervals
containing ants. The contents of the four parts are stated as below.

\subsection{Partition of Interval and Initialization}

Suppose interval $[a,b]$ is partitioned into $n$\ small intervals with equal
size, which are denoted by $\{I_{1},I_{2},\cdots ,I_{n}\}$, where $n$ is a
pre-assigned number. Then each interval has size

\begin{equation*}
\delta =\frac{b-a}{n}
\end{equation*}

\begin{equation*}
I_{i}=[a+(i-1)\delta ,a+i\delta ]
\end{equation*}

The $i-th$\ interval $I_{i}$\ has center $x_{i}$.

\begin{equation*}
x_{i}=a+(i-\frac{1}{2})\delta
\end{equation*}

Suppose $t$ denotes the $t-th$ iteration step of ACO and it is initialized
as zero (i.e., $t=0$). Put $n$ ants at the centers of the $n$\ intervals,
and each interval has only one ant. Suppose these ants are denoted by $%
a_{1},a_{2},\cdots a_{n}$ respectively, and ant $a_{i}$ is associated with
the $i-th$\ interval $I_{i}$. Each ant will release an initial pheromone at
its associated interval $I_{i}$ (i.e., $\tau _{i}(0)=const$, $const$ is a
constant number). In addition, set the increment of the pheromone of each
interval to zero (i.e., $\Delta \tau _{i}(0)=0$)

Figure 1 shows a diagram of initialization.

\begin{figure}[tbh]
\epsfig{file=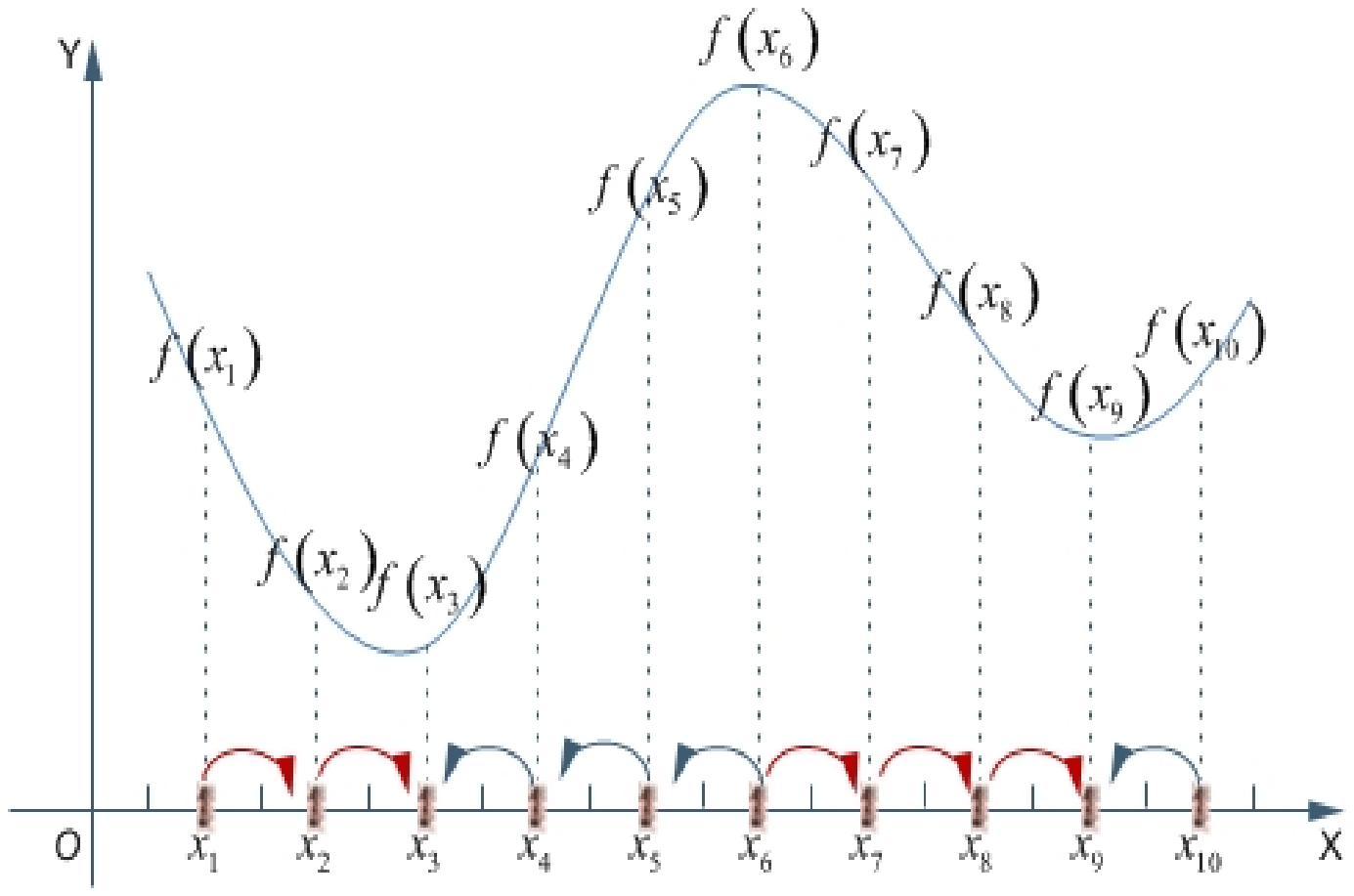,width=10cm,}
\caption{Initialization: Divide the function domain into intervals and put
an ant at the center of every interval.}
\end{figure}

\subsection{Rule of Ant Moving}

Let $Neighbor(I_{i})$ be a set of neighboring intervals of $I_{i}$. Take
one-dimensional function for example

\begin{equation*}
Neighbor(I_{i})=\left\{
\begin{array}{l}
\left\{ I_{i+1}\right\} ,\text{ \ \ \ \ \ \ \ \ }i=1 \\
\left\{ I_{i-1},I_{i+1}\right\} ,\text{ \ \ }i=2\cdots n-1 \\
\left\{ I_{i-1}\right\} ,\text{ \ \ \ \ \ \ \ \ }i=n%
\end{array}%
\right.
\end{equation*}

As it is discussed in section $A$, the ant staying at interval $I_{i}$ will
move to neighbor interval denoted by $I_{j}$, just like there is a virtual
edge $e(I_{i},I_{j})$. The weight of the virtual edge is $\left\vert
f(x_{i})-f(x_{j})\right\vert $. Then the heuristic factor is

\begin{equation*}
\eta _{ij}=\left\vert f(x_{i})-f(x_{j})\right\vert
\end{equation*}

Suppose interval $I_{i}$ contains ant $a_{k}$. If $f(x_{i})>f(x_{j})$, ant $%
a_{k}$ is allowed to move to neighbor interval $I_{j}$. Otherwise, it is
forbidden to move. Suppose all intervals allowed to be accessed by ant $%
a_{k} $\ is marked as $allowed_{k}$. The transition probability of ant $%
a_{k} $\ is defined as

\begin{equation}
p_{ij}^{(k)}(t)=\left\{
\begin{tabular}{ccc}
$\frac{\tau _{j}^{\alpha }(t).\eta _{ij}^{\beta }}{\underset{h\in allowed_{k}%
}{\sum }\tau _{h}^{\alpha }(t).\eta _{ih}^{\beta }},$ & $\ \ \ if$ & $%
f(x_{i})>f(x_{j})$ \\
$0,$ & \multicolumn{2}{c}{$otherwise$}%
\end{tabular}%
\right.  \label{Function probability}
\end{equation}

In Eq.\ref{Function probability}, $\alpha $ is the relative influence of the
trail strength; $\beta $\ is the heuristic information; $\tau _{j}(t)$ is
the pheromone of interval $I_{j}$.

\subsection{The Rule of Pheromone Updating}

Suppose $a_{k}$ ant is staying at interval $I_{i}$ and it will move to
neighbor interval $I_{j}$. After it moves to $I_{j}$, releases pheromone at $%
I_{j}$. The pheromone amount is denoted by $\Delta \tau _{j}^{(k)}(t)$

\begin{equation*}
\Delta \tau _{j}^{(k)}(t)=C_{1}(f(x_{i})-f(x_{j}))
\end{equation*}

, where $C_{1}$ is a positive constant.

The bigger $f(x_{i})-f(x_{j})$ is, the higher the amount of released
pheromone is, the more possibly that other ants will be attracted to
interval $I_{j}$.

Not only ant $a_{k}$ arrives $I_{j}$ and releases pheromone, but also other
ants which move to interval $I_{j}$ and release pheromone too. Suppose there
are $q$ ants will move to interval $I_{j}$ during the $t-th$ iteration step,
which are denoted by $a_{j_{1}},a_{j_{2}},\cdots ,a_{j_{q}}$. The sum of
released pheromone by $a_{j_{1}},a_{j_{2}},\cdots ,a_{j_{q}}$ is $\Delta
\tau _{j}(t)$. Then

\begin{equation*}
\Delta \tau _{j}(t)=\sum_{p=1}^{q}\Delta \tau _{j}^{(j_{p})}(t)
\end{equation*}

When all ants $a_{j_{1}},a_{j_{2}},\cdots ,a_{j_{q}}$\ move to $I_{j}$, the
amount of pheromone $\tau _{j}$ is changed as Eq.\ref{Function update}.

\begin{equation}
\tau _{j}(t+1)=(1-\rho )\cdot \tau _{j}(t)+\Delta \tau _{j}(t)
\label{Function update}
\end{equation}

, where $\rho $ is the evaporation percentage of the trail (thus, $1-\rho $\
corresponds to persistence).

\subsection{Keeping Only the Intervals Containing Ants to Cut Down Search
Range}

The intervals that have smaller function values depositing much more
pheromone, and it will attract ants more powerfully. After several
iterations, the distribution of ants has the feature that all ants stay at
the intervals that have smaller function values and other intervals contain
no ants. That is, extreme points are included in the intervals containing
ants. And then keep the intervals having ant and delete other intervals to
update search range. Thus, the updated search range became smaller. Dividing
the updated search range into smaller intervals will result in much smaller
search range at next iteration step. When intervals becoming sufficient
small, all ants will stay around extreme points, the centers of the
intervals are their approximations.

Figure 2 shows the distribution feature of ants.

In addition, as it is well known, ACO runs slow, which is the
bottleneck of application. And it evades this bottleneck that
keeping only the intervals having ants to cut down search range.

\begin{figure}[tbh]
\epsfig{file=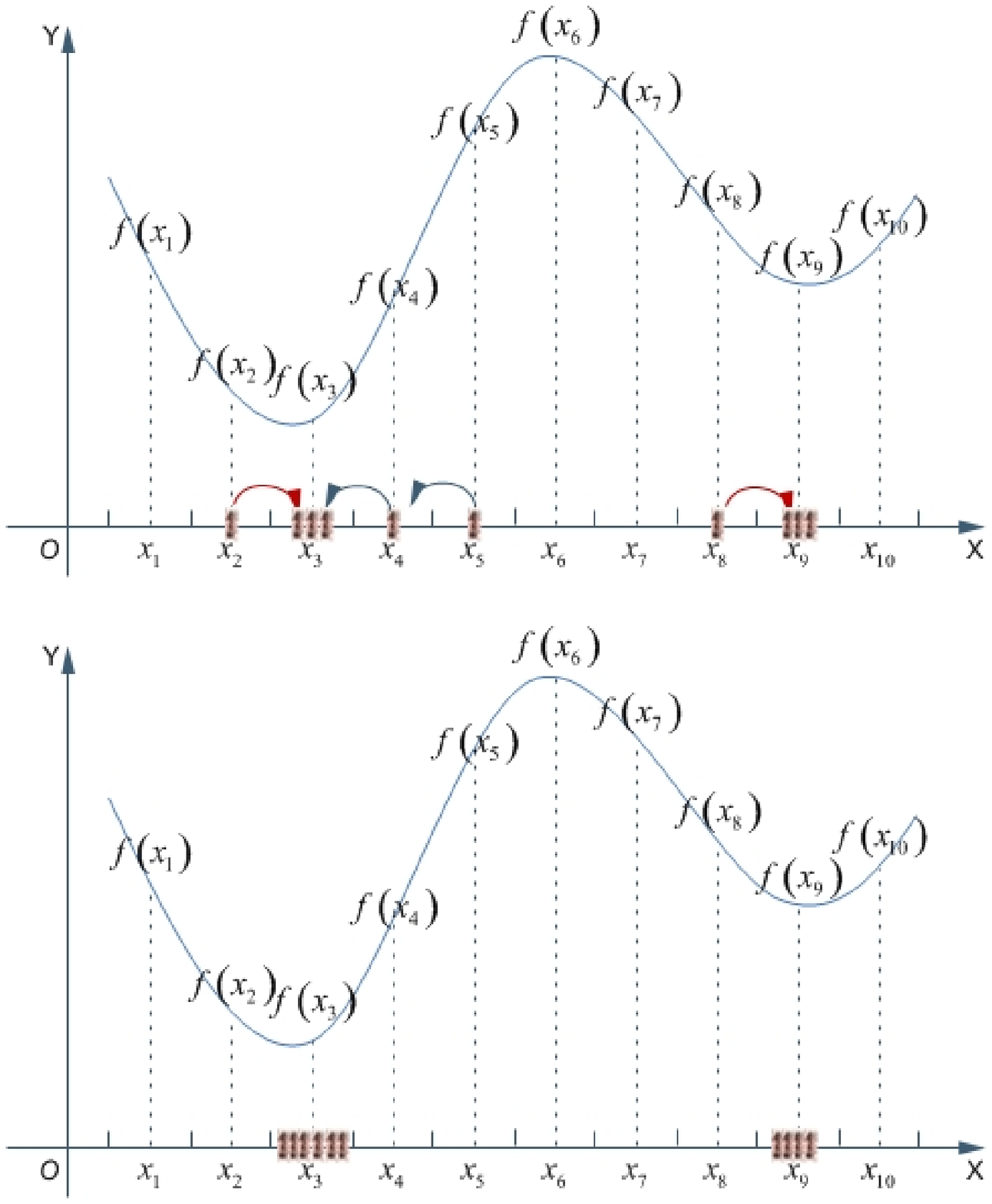,width=10cm,}
\caption{The Feature of Ant Distribution: The intervals including extreme
points have more pheromone to attract ants. After several iteration steps,
all ants will stay at the intervals containing extreme points while other
intervals are empty. Then search range become smaller, which consists of the
intervals containing ants. At last all ants stay around extreme points.}
\end{figure}

\subsection{Method of Searching All ExtremePoints}

\textbf{Step1 (Initialization):} Divide domain $[a,b]$ into many small
intervals and put an ant in each interval; Do other initialization. The
detail is shown at section $B$. Suppose $\delta $ is the length of interval
and $\varepsilon $ is a stop threshold.

\textbf{Step2} While ($\delta >\varepsilon $)

\{

\ \ \textbf{Step2.1:} All ants move to new intervals according the rule
shown at section $C$

\ \ \textbf{Step2.2:} Update pheromone according the rule shown at section $%
D $

\ \ \textbf{Step2.3:} Update search range according to section $E$ and
divide it into smaller intervals (Suppose the number of these intervals is $%
n1$) . Calculate the size of interval and set it to $\delta $.

\}

\textbf{Step3} Extract all intervals that contain ants, the centers of the
intervals are the approximations of extreme points.

If argument $x$ is multi-dimensional vector, divide the range of every
component of vector into smaller intervals, the combination of these
intervals forms many small lattices. And then put an ant in each lattice,
apply the above method, all extreme points can be extracted.

\subsection{An Example}

To understand above method easily, an simple example is stated as below:

Assume that the domain of 1-dimensional function is divided into 3 intervals
$I_{1}$, $I_{2}$, $I_{3}$, which associated center is $x_{1}$, $x_{2}$ and $%
x_{3}$ respectively. Initially ant $a_{1}$, $a_{2}$, and $a_{3}$ is put at $%
x_{1}$, $x_{2}$ and $x_{3}$ respectively.

Check the first ant: If $f(x_{1})>f(x_{2})$, ant $a_{1}$ moves to interval $%
I_{2}$. Otherwise, do nothing.

Check the 2nd ant: If their is unique interval (e.g. $I_{3}$ ) such that $%
f(x_{2})>f(x_{3})$, ant $a_{2}$ moves to interval $I_{3}$. If $f(x_{2})$ is
smaller than both $f(x_{1})$ and $f(x_{3})$, do nothing. If $f(x_{2})$ is
bigger than both $f(x_{1})$ and $f(x_{3})$, it is uncertain that ant $a_{2}$
moves to $I_{1}$ or $I_{3}$. And ant $a_{2}$ will select its visiting
interval randomly according to its transition probability defined at Eq.\ref%
{Function probability}.

Check the 3rd ant using same way.

After all ants are checked, update their associated interval
(position) and interval pheromone. Repeat above processing until all
ants can not move.

Then keep the intervals which contains ants, and delete other blank
intervals. And divided the intervals containing ants into smaller interval,
repeat above process until the size of interval is sufficient small. And
then all interval centers are the approximations of extreme points.

\section{Experiment}

In this section, several functions will be tested. The parameters are listed
as below:

$const=10$, $\alpha =1$, $\beta =1$, $C_{1}=1$, $\rho =0.3$, $\varepsilon
=0.0001$

Two performances are considered, which are error (ratio of inaccuracy) $r$\
and runtime. Error $r$ is defined as

\begin{equation*}
r=\left( \left\vert \frac{f(x_{0}^{\prime })-f(x_{0})}{f(x_{0})}\right\vert
\right) \times 100\%
\end{equation*}

, where $(x_{0},f(x_{0}))$ denotes the true extreme point on theory and $%
(x_{0}^{\prime },f(x_{0}^{\prime }))$ is its approximation calculated by the
method presented in this paper.

In addition, the hardware condition is: notebook PC DELL D520, CPU 1.66 GHZ.

\begin{figure}[tbh]
\epsfig{file=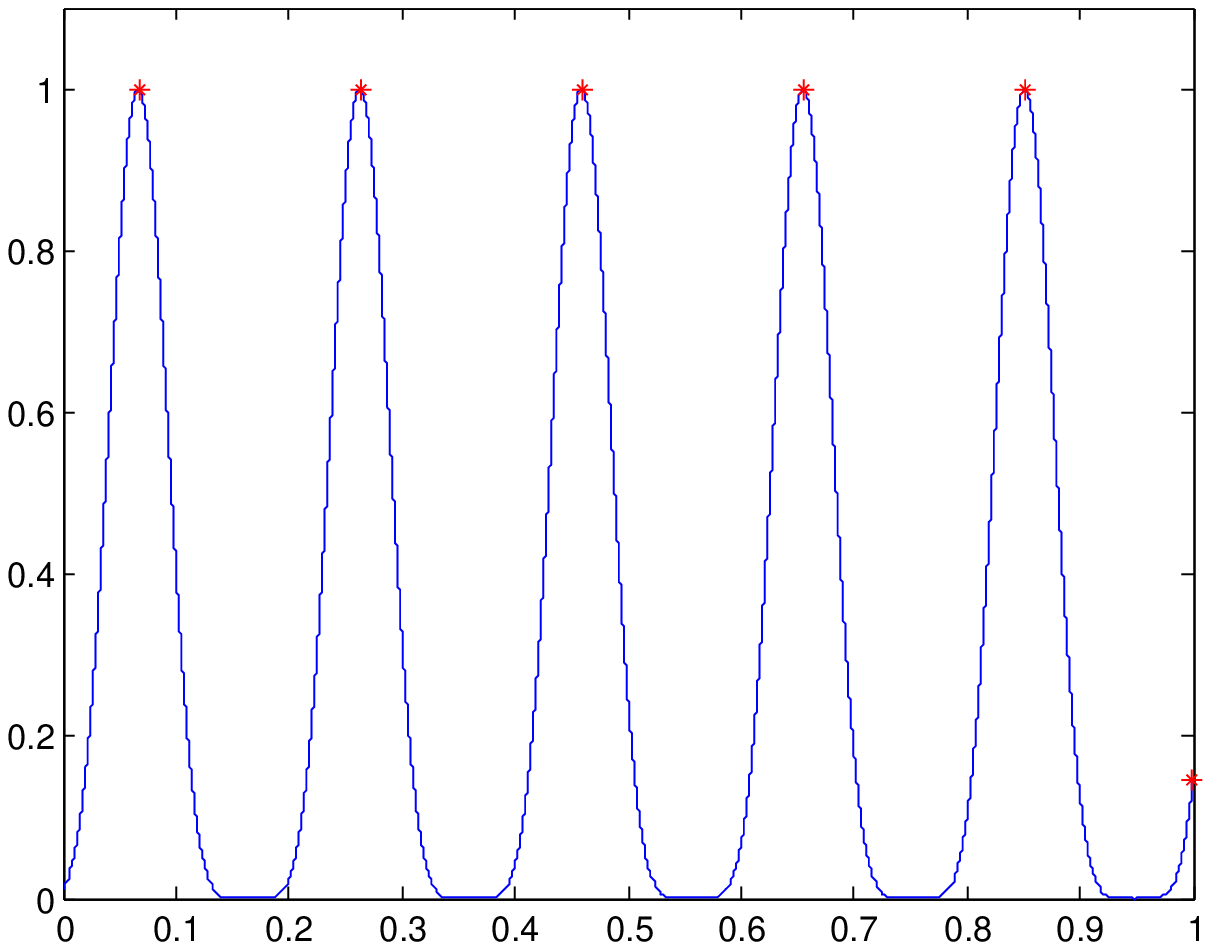,width=10cm,}
\caption{{}\textbf{The  Extreme Points  Calculated of Function }%
{\protect\small \ }$f_{1}(x)$\textbf{. }}
\label{figInstance1}
\end{figure}

\subsection{Instance 1 (see table 1 and Fig.\protect\ref{figInstance1}):}

$f_{1}(x)=\sin ^{6}(5.1\pi x+0.5),\ \ x\in \lbrack 0,1]$

In the experiment, additional parameter is $n=20$ and $n1=10$. Table 1 and
Fig.\ref{figInstance1} show all extreme points (local maximal points) of $%
f_{1}(x)$. Only 0.7106 seconds is cost.

Other functions are tested, their errors are less than $10^{-8}$ except the
boundary. And runtime is less than 1 second (see appendix I).

\subsection{Instance 2 (see Fig.\protect\ref{figInstance2}):}

$f_{2}(x)=5e^{-0.5x}\sin (30x)+e^{0.2x}\sin (20x)+6,\ \ x\in \lbrack 0,8]$

Instance 2 is a typical test function, which include many extreme points and
any small change of argument $x$ will result in big change. In addition, the
theoretical calculation of extreme points of instance 2 is difficult.

The additional parameters are $n=480$ and $n1=10$. Fig.\ref{figInstance2}
shows all the calculated extreme points, and the real numbers are listed at
appendix (see appendix II).

\subsection{Instance 3 (see Fig.\protect\ref{figInstance3}):}

$f_{3}(x_{1},x_{2})=x_{1}^{2}+x_{2}^{2}-\cos (18x_{1})-\cos (18x_{2})$, \ $%
x_{1},x_{2}\in \lbrack -1,1]$

The interval $[-1,1]\otimes \lbrack -1,1]$ is divided into $n=40\times 40$
intervals initially. And at next iteration steps, search domain is divided
into $n1=20\times 20$ small intervals. 36 extreme points are got and shown
at Fig.\ref{figInstance3}, and the digital solutions are listed at appendix
(see appendix III)

Instance 3 is a 2-dimensional function and 3203.2968 seconds is cost. And it
is slower than 1-dimensional function instance 1 and instance 2. To improve
running speed is the next work.

Many functions are tested by the authors, and some experiments
results are listed at appendix. These tests demonstrate that
solution error is less than $10^{-8}$ except the special case that
extreme point is at the border of the domain. In addition, these
testes also demonstrate that the method is very fast for
1-dimensional function.

\bigskip

\begin{tabular}{|c|c|c|c|}
\hline
\multicolumn{4}{|c|}{\textbf{TABLE 1. The Extreme Points of Function }$%
f_{1}(x)$} \\ \hline
{\small Extreme} & {\small Theory Value} & {\small Calculated Value} &
{\small Error (\%)} \\ \hline
{\small Points} & $\left( x_{0},f(x_{0})\right) $ & $\left( x_{0}^{\prime
},f(x_{0}^{\prime })\right) $ & ${\small r}$ \\ \hline
{\small 1} & $%
\begin{array}{c}
\text{{\small 0.066832364,}} \\
\text{{\small 1}}%
\end{array}%
$ & $%
\begin{array}{c}
\text{{\small 0.066827175,}} \\
\text{{\small 0.999999979}}%
\end{array}%
$ & {\small 2.1e-06} \\ \hline
{\small 2} & $%
\begin{array}{c}
\text{{\small 0.262910795,}} \\
\text{{\small 1}}%
\end{array}%
$ & $%
\begin{array}{c}
\text{{\small 0.262914975,}} \\
\text{{\small 0.999999987}}%
\end{array}%
$ & {\small 1.3e-06} \\ \hline
{\small 3} & $%
\begin{array}{c}
\text{{\small 0.458989227,}} \\
\text{{\small 1}}%
\end{array}%
$ & $%
\begin{array}{c}
\text{{\small 0.458990475,}} \\
\text{{\small 0.999999999}}%
\end{array}%
$ & {\small 1.0e-07} \\ \hline
{\small 4} & $%
\begin{array}{c}
\text{{\small 0.655067658,}} \\
\text{{\small 1}}%
\end{array}%
$ & $%
\begin{array}{c}
\text{{\small 0.655066175,}} \\
\text{{\small 0.999999998}}%
\end{array}%
$ & {\small 2.0e-07} \\ \hline
{\small 5} & $%
\begin{array}{c}
\text{{\small 0.851146090,}} \\
\text{{\small 1}}%
\end{array}%
$ & $%
\begin{array}{c}
\text{{\small 0.851148550,}} \\
\text{{\small 0.999999995}}%
\end{array}%
$ & {\small 5.0e-07} \\ \hline
{\small 6} & $%
\begin{array}{c}
\text{{\small 1}} \\
\text{{\small 0.147822118}}%
\end{array}%
$ & $%
\begin{array}{c}
\text{{\small 0.999998400,}} \\
\text{{\small 0.147800654}}%
\end{array}%
$ & {\small 0.0145} \\ \hline
\multicolumn{4}{|c|}{%
\begin{tabular}{l}
{\small Conclusion: Error of the method presented this paper is low except
the border point ( 6th point). } \\
{\small And runtime is fast (0.7106s). Notice: The 6 points are shown at Fig.%
\ref{figInstance1}}%
\end{tabular}%
{\small \ }} \\ \hline
\end{tabular}

Notice: From the Table 1, we can see that the border point has big error
because the value calculated is the center\ of the interval, not boundary.
To evade this drawback, the function value at boundary can be calculated
directly.

\begin{figure}[tbh]
\epsfig{file=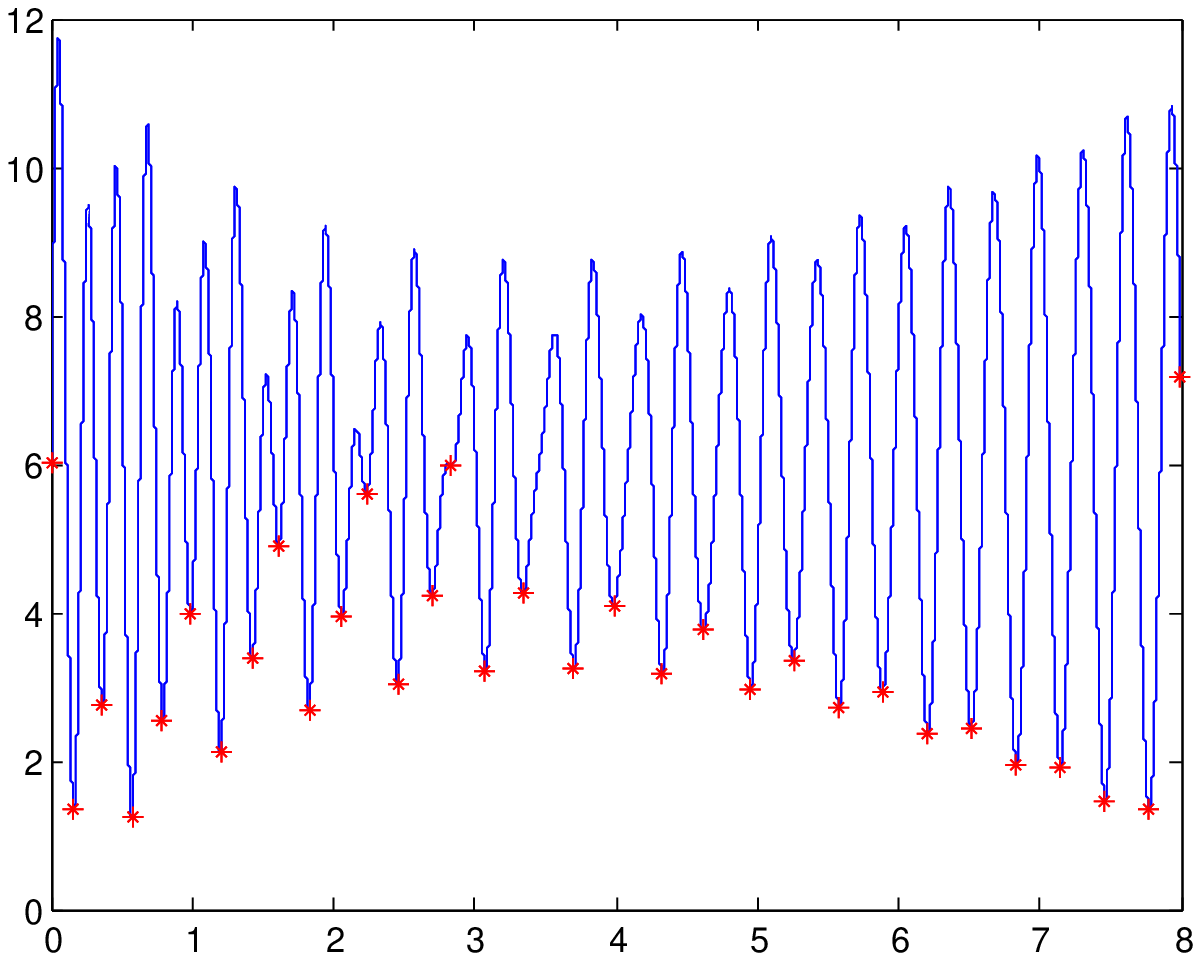,width=10cm,}
\caption{\textbf{The  Extreme Points Calculated of Function }$f_{2}(x)$}
\label{figInstance2}
\end{figure}

\begin{figure}[tbh]
\epsfig{file=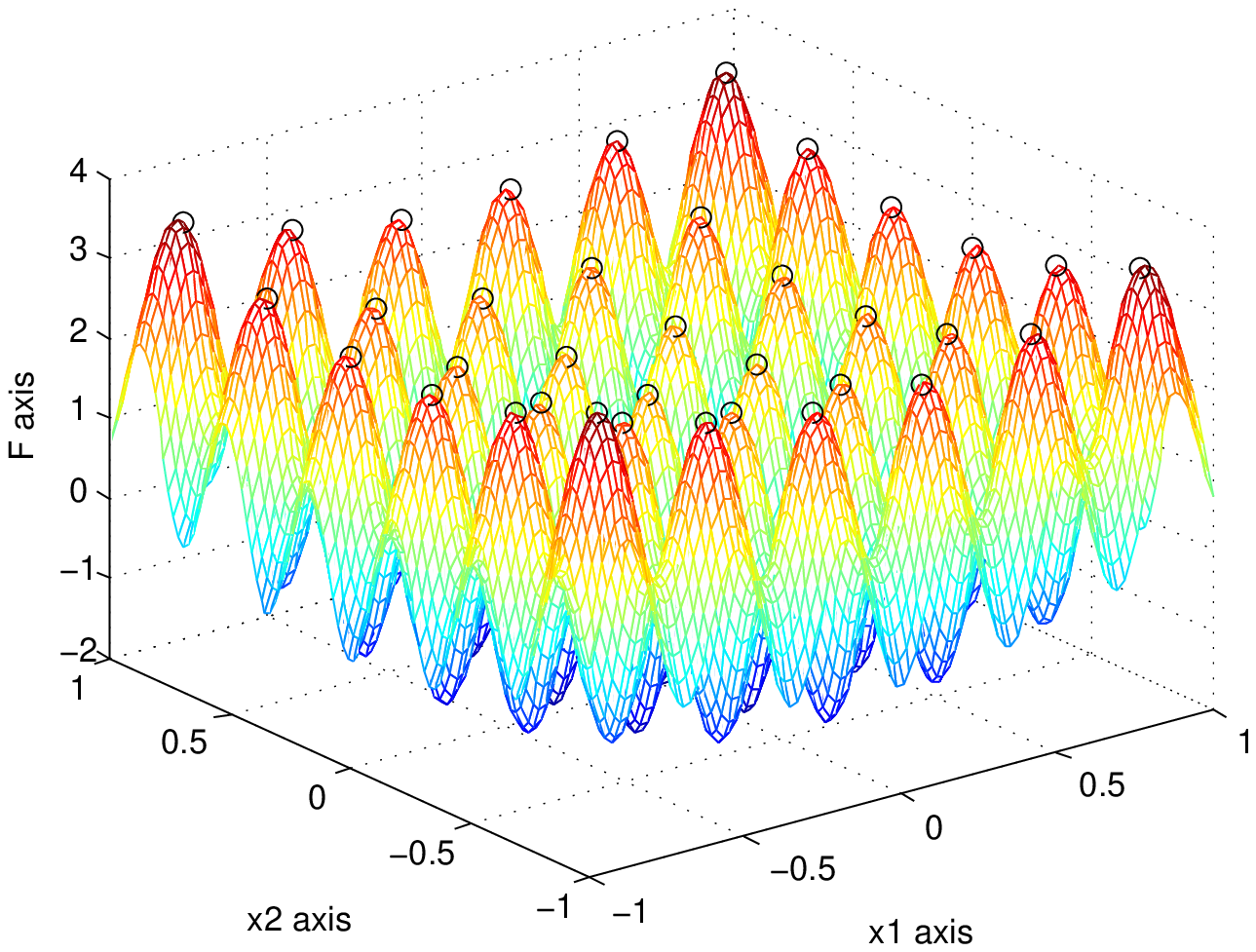,width=10cm,}
\caption{\textbf{The Extreme Points Calculated of Function }$%
f_{3}(x_{1},x_{2})$\textbf{.}}
\label{figInstance3}
\end{figure}

\section{Conclusion}

To find all extreme points of multimodal functions is called extremum
problem, which is a well known difficult issue in optimization fields. It is
reported rarely that applying Ant Colony Optimization(ACO) to solve the
problem. And the motivation of this paper is to explore ACO application
method to solve it. In this paper, the following method is presented:

Divide the domain of function into many intervals and put an ant in each
interval. And then design rule such that every ant moves to the interval
containing extreme point near by. At last all ants stay around extreme
points.

The method presented in this paper has following three advantages:

1. Solution accuracy is high. Experiment shows that solution error
is less than $10^{-8}$.

2. Solution calculated is stable (robust). Ant only indicates the interval
containing extreme point, not the accurate position of extreme point. It is
easy for ant to find a interval although finding a special point in interval
is difficult.

3. The method is fast for 1-dimensional function. ACO is slow. But some
feature is found to speed ACO (see section 2.5)

\begin{acknowledgments}
The authors appreciate the discussion from the members of Gene Computation
Group, J. Gang, X. Li, C.-B. Wang, W. Hu, S.-P. Wang, Q. Yang, J.-L. Zhou,
P. Shuai, L.-J. Ye. The authors appreciate the help from Prof. J. Zhang,
Z.-Lin Pu, X.-P. Wang, J. Zhou, and Q. Li.
\end{acknowledgments}

\bigskip

\section{Appendix}

\subsection{Appendix I}

\subsubsection{\textbf{Instance 4}:}

$f_{4}(x)=(x+1)(x+2)(x+3)(x+4)(x+5)+5$ \ \ \ \ \ $x\in \lbrack -5,0]$

The additional parameters are $n=30$ and \ $n_{1}=10$.

\bigskip

\begin{tabular}{|c|c|c|c|}
\hline
\multicolumn{4}{|l|}{\ \ \ \ \ \ \ \ \ \ \ \ \ \textbf{Table 2. Extreme
Points of Function }$f_{4}(x)$} \\ \hline
&  &  &  \\ \hline
{\small Extreme} & {\small Theory Value} & {\small Calculated Value} &
{\small Error (\%)} \\ \hline
{\small Points} & $\left( x_{0},f(x_{0})\right) $ & $\left( x_{0}^{\prime
},f(x_{0}^{\prime })\right) $ & ${\small r}$ \\ \hline
{\small 1} & $%
\begin{array}{c}
{\small -5.00000000000000,} \\
{\small 5.00000000000000}%
\end{array}%
$ & $%
\begin{array}{c}
{\small -4.99999893333333,} \\
{\small 5.00002559994310}%
\end{array}%
$ & {\small 5.12e-004} \\ \hline
{\small 2} & $%
\begin{array}{c}
{\small -3.54391225590234,} \\
{\small 3.58130337441708}%
\end{array}%
$ & $%
\begin{array}{c}
{\small -3.54391549166667,} \\
{\small 3.58130337448565}%
\end{array}%
$ & {\small 1.91e-009} \\ \hline
{\small 3} & $%
\begin{array}{c}
{\small -1.35556713184173,} \\
{\small 1.36856779155116}%
\end{array}%
$ & $%
\begin{array}{c}
{\small -1.35557000833333,} \\
{\small 1.36856779171500}%
\end{array}%
$ & {\small 1.20e-008} \\ \hline
\multicolumn{4}{|c|}{\small Notice: 1. Runtime is 0.5280s. 2. From the
table, we can see that only the boundary} \\ \hline
\multicolumn{4}{|l|}{{\small maximum value has big error because }$%
x_{0}^{\prime }${\small \ is the center of interval, not the boundary.}} \\
\hline
\multicolumn{4}{|l|}{\small To evade this drawback,the function value at
boundary can be calculated directly.} \\ \hline
\multicolumn{4}{|l|}{} \\ \hline
\end{tabular}

\bigskip

\begin{figure}[tbh]
\epsfig{file=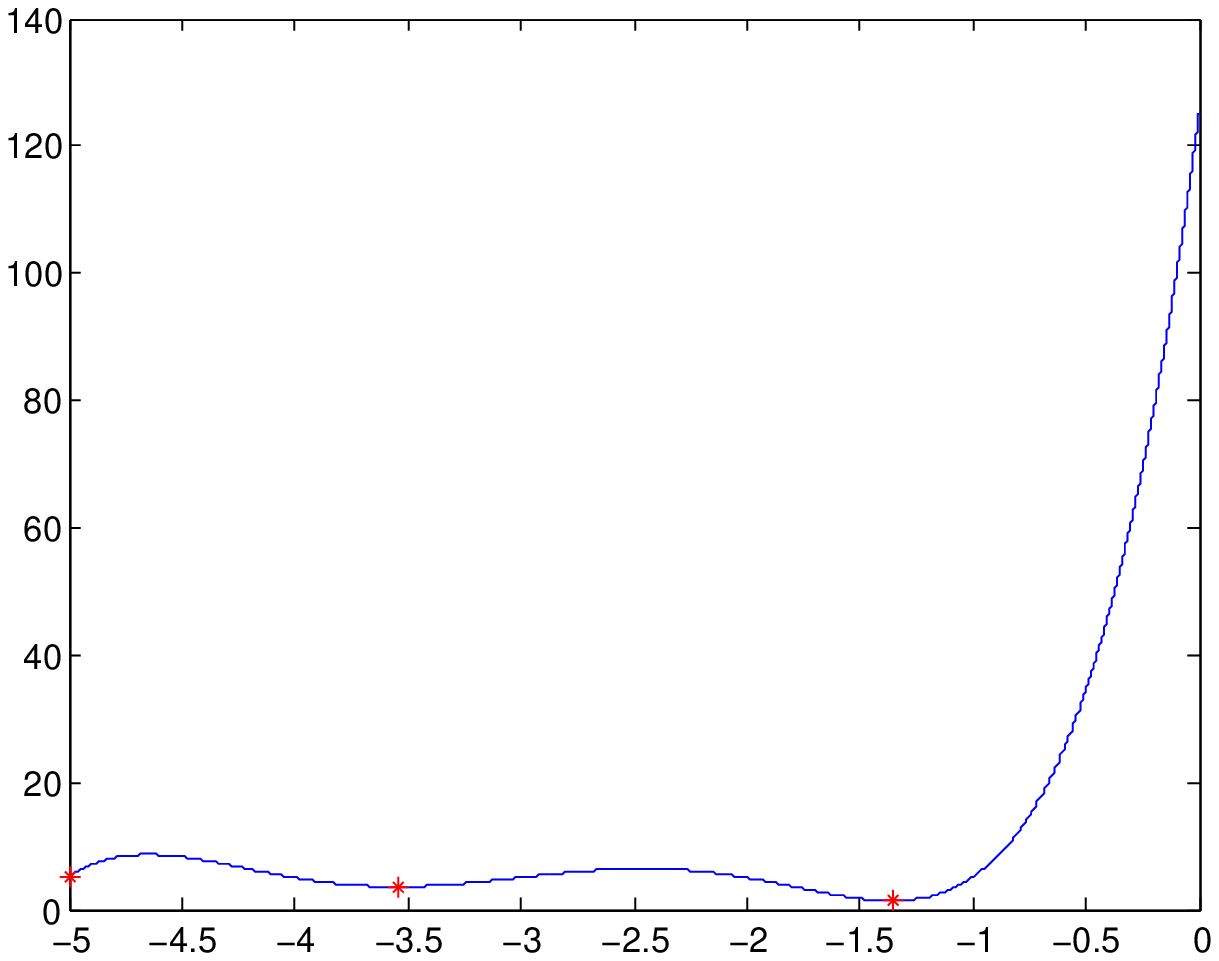,width=10cm,}
\caption{\textbf{The  Extreme Points Calculated of Function }$f_{4}(x)$%
\textbf{.} The red stars are the extreme points.}
\end{figure}

\subsubsection{\textbf{Instance 5:}}

$f_{5}(x)=(x+2)\cos (9x)+\sin (7x)$ \ \ \ $x\in \lbrack 0,4]$

The additional parameters are $n=95$ and $n_{1}=10$

\bigskip

\
\begin{tabular}{|c|c|c|c|}
\hline
\multicolumn{4}{|l|}{\ \ \ \ \ \ \ \ \ \ \ \ \textbf{Table 3. Extreme Points
of Function }$f_{5}(x)$} \\ \hline
&  &  &  \\ \hline
{\small Extreme} & {\small Theory Value} & {\small Calculated Value} &
{\small Error (\%)} \\ \hline
{\small Points} & $\left( x_{0},f(x_{0})\right) $ & $\left( x_{0}^{\prime
},f(x_{0}^{\prime })\right) $ & ${\small r}$ \\ \hline
{\small 1} & $%
\begin{array}{c}
{\small 0.0000000000000,} \\
{\small 2.0000000000000}%
\end{array}%
$ & $%
\begin{array}{c}
{\small 0.00000134736842,} \\
{\small 2.00001077880032}%
\end{array}%
$ & {\small 5.39e-004} \\ \hline
{\small 2} & $%
\begin{array}{c}
{\small 0.38762423574739,} \\
{\small -1.8300394960224}%
\end{array}%
$ & $%
\begin{array}{c}
{\small 0.38762006315789,} \\
{\small -1.83003949456305}%
\end{array}%
$ & {\small 7.97e-008} \\ \hline
{\small 3} & $%
\begin{array}{c}
{\small 1.03491254831578,} \\
{\small -2.19650045294125}%
\end{array}%
$ & $%
\begin{array}{c}
{\small -1.03490865263158,} \\
{\small -2.19650045140718}%
\end{array}%
$ & {\small 16.98e-008} \\ \hline
{\small 4} & $%
\begin{array}{c}
{\small 1.72787330641709,} \\
{\small -4.13597012121530}%
\end{array}%
$ & $%
\begin{array}{c}
{\small 1.72787490526316,} \\
{\small -4.13597012080927}%
\end{array}%
$ & {\small 9.82e-009} \\ \hline
{\small 5} & $%
\begin{array}{c}
{\small 2.44888001781347,} \\
{\small -5.43427465397202}%
\end{array}%
$ & $%
\begin{array}{c}
{\small 2.44888418947368,} \\
{\small -5.43427465041009}%
\end{array}%
$ & {\small 6.55e-008} \\ \hline
{\small 6} & $%
\begin{array}{c}
{\small 3.16064032530539,} \\
{\small -5.21793460286144}%
\end{array}%
$ & $%
\begin{array}{c}
{\small 3.16064037894737,} \\
{\small -5.21793460286083}%
\end{array}%
$ & {\small 1.17e-011} \\ \hline
{\small 7} & $%
\begin{array}{c}
{\small 3.84493263713282,} \\
{\small -4.86068863199737}%
\end{array}%
$ & $%
\begin{array}{c}
{\small 3.84493433684211,} \\
{\small -4.86068863138223}%
\end{array}%
$ & {\small 1.27e-008} \\ \hline
\multicolumn{4}{|c|}{\small Notice: 1. The runtime is 0.9030s. 2. The
boundary maximum value has big error \ \ } \\ \hline
\multicolumn{4}{|c|}{{\small because }$x_{0}^{\prime }$ {\small is the
center of the interval, not the boundary. To evade this drawback, }} \\
\hline
\multicolumn{4}{|l|}{\small the function value at boundary can be calculated
directly. \ \ \ \ \ \ \ \ \ \ \ \ \ \ \ \ \ \ \ \ \ \ \ } \\ \hline
\multicolumn{4}{|l|}{} \\ \hline
\end{tabular}

\begin{figure}[tbh]
\epsfig{file=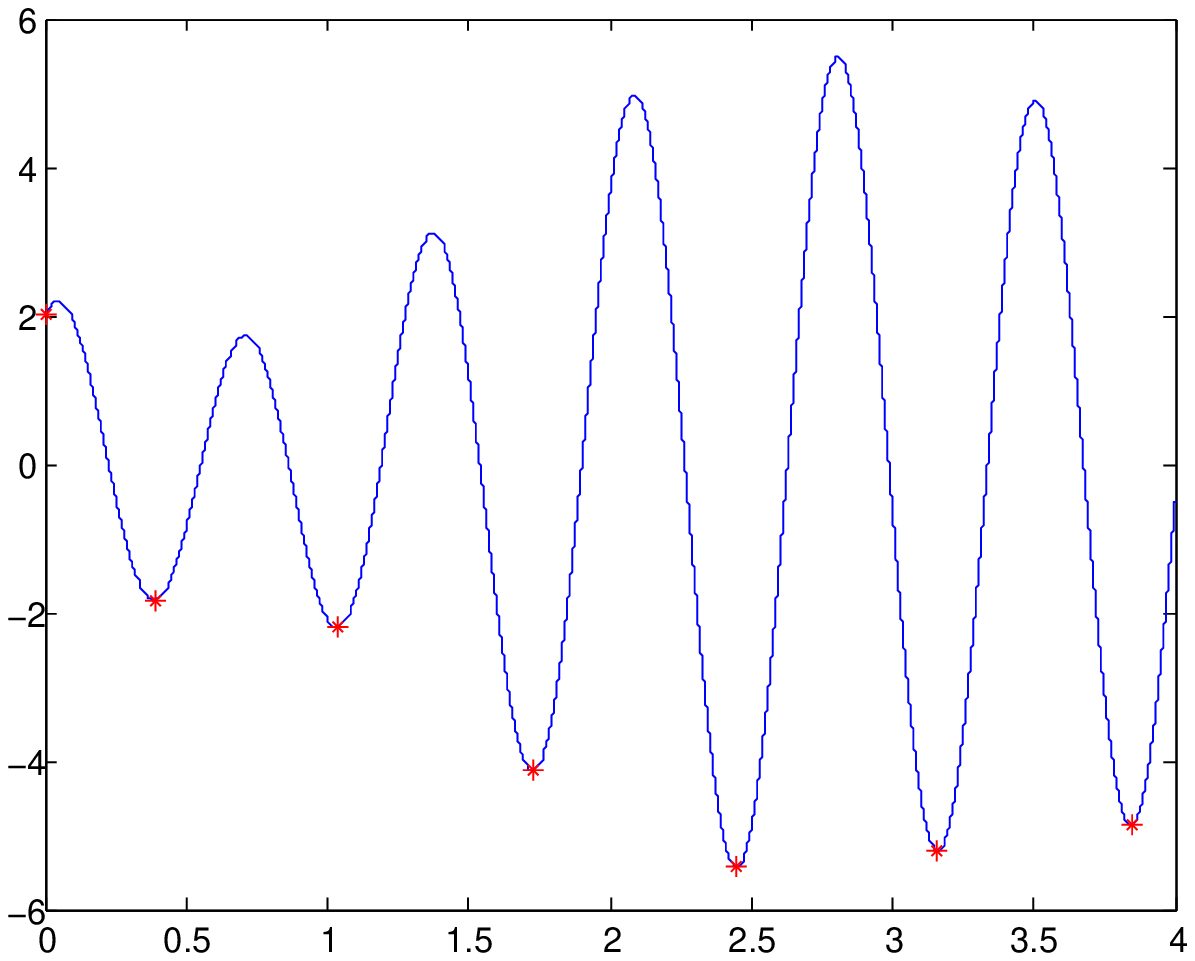,width=10cm,}
\caption{\textbf{The  Extreme Points Calculated of Function  }$f_{5}(x)$}
\end{figure}

\subsection{Appendix II.}

$f_{2}(x)=5e^{-0.5x}\sin (30x)+e^{0.2x}\sin (20x)+6,$ $\ \ \ \ x\in \lbrack
0,8]$

The additional parameters are $n=480$ and $n1=10$.

\bigskip

\begin{tabular}{|c|c|c|c|c|c|}
\hline
\multicolumn{6}{|l|}{\ \ \ \ \ \ \ \ \ \textbf{Table 4. Extreme Points of
Function }$f_{2}(x)$} \\ \hline
\multicolumn{6}{|c|}{} \\ \hline
{\small Extreme} & {\small Calculated Value} & {\small Extreme} & {\small %
Calculated Value} & {\small Extreme} & {\small Calculated Value} \\ \hline
{\small Points} & $\left( x_{0}^{\prime },f(x_{0}^{\prime })\right) $ &
{\small Points} & $\left( x_{0}^{\prime },f(x_{0}^{\prime })\right) $ &
{\small Points} & $\left( x_{0}^{\prime },f(x_{0}^{\prime })\right) $ \\
\hline
{\small 1} & $%
\begin{array}{c}
{\small 0.0000,} \\
{\small 6.0005}%
\end{array}%
$ & {\small 12} & $%
\begin{array}{c}
{\small 2.2337,} \\
{\small 5.5878}%
\end{array}%
$ & {\small 23} & $%
\begin{array}{c}
{\small 5.2547,} \\
{\small 3.3639}%
\end{array}%
$ \\ \hline
{\small 2} & $%
\begin{array}{c}
{\small 0.1615,} \\
{\small 1.3372}%
\end{array}%
$ & {\small 13} & $%
\begin{array}{c}
{\small 2.4521,} \\
{\small 3.0483}%
\end{array}%
$ & {\small 24} & $%
\begin{array}{c}
{\small 5.5812,} \\
{\small 2.7148}%
\end{array}%
$ \\ \hline
{\small 3} & $%
\begin{array}{c}
{\small 0.3627,} \\
{\small 2.7441}%
\end{array}%
$ & {\small 14} & $%
\begin{array}{c}
{\small 2.6982,} \\
{\small 4.2229}%
\end{array}%
$ & {\small 25} & $%
\begin{array}{c}
{\small 5.8862,} \\
{\small 2.9274}%
\end{array}%
$ \\ \hline
{\small 4} & $%
\begin{array}{c}
{\small 0.5725,} \\
{\small 1.2573}%
\end{array}%
$ & {\small 15} & $%
\begin{array}{c}
{\small 2.8368,} \\
{\small 5.9928}%
\end{array}%
$ & {\small 26} & $%
\begin{array}{c}
{\small 6.2081,} \\
{\small 2.3729}%
\end{array}%
$ \\ \hline
{\small 5} & $%
\begin{array}{c}
{\small 0.7926,} \\
{\small 2.5460}%
\end{array}%
$ & {\small 16} & $%
\begin{array}{c}
{\small 3.0777,} \\
{\small 3.2186}%
\end{array}%
$ & {\small 27} & $%
\begin{array}{c}
{\small 6.5164,} \\
{\small 2.4487}%
\end{array}%
$ \\ \hline
{\small 6} & $%
\begin{array}{c}
{\small 0.9889,} \\
{\small 3.9745}%
\end{array}%
$ & {\small 17} & $%
\begin{array}{c}
{\small 3.3398,} \\
{\small 4.2595}%
\end{array}%
$ & {\small 28} & $%
\begin{array}{c}
{\small 6.8355,} \\
{\small 1.9567}%
\end{array}%
$ \\ \hline
{\small 7} & $%
\begin{array}{c}
{\small 1.1995,} \\
{\small 2.1270}%
\end{array}%
$ & {\small 18} & $%
\begin{array}{c}
{\small 3.7032,} \\
{\small 3.2478}%
\end{array}%
$ & {\small 29} & $%
\begin{array}{c}
{\small 7.1485,} \\
{\small 1.9215}%
\end{array}%
$ \\ \hline
{\small 8} & $%
\begin{array}{c}
{\small 1.4252,} \\
{\small 3.3895}%
\end{array}%
$ & {\small 19} & $%
\begin{array}{c}
{\small 3.9822,} \\
{\small 4.0777}%
\end{array}%
$ & {\small 30} & $%
\begin{array}{c}
{\small 7.4631,} \\
{\small 1.4650}%
\end{array}%
$ \\ \hline
{\small 9} & $%
\begin{array}{c}
{\small 1.6133,} \\
{\small 4.9025}%
\end{array}%
$ & {\small 20} & $%
\begin{array}{c}
{\small 4.3288,} \\
{\small 3.1621}%
\end{array}%
$ & {\small 31} & $%
\begin{array}{c}
{\small 7.7748,} \\
{\small 1.3366}%
\end{array}%
$ \\ \hline
{\small 10} & $%
\begin{array}{c}
{\small 1.8261,} \\
{\small 2.7009}%
\end{array}%
$ & {\small 21} & $%
\begin{array}{c}
{\small 4.6206,} \\
{\small 3.7560}%
\end{array}%
$ & {\small 32} & $%
\begin{array}{c}
{\small 8.0000,} \\
{\small 7.1737}%
\end{array}%
$ \\ \hline
{\small 11} & $%
\begin{array}{c}
{\small 2.0601,} \\
{\small 3.9371}%
\end{array}%
$ & {\small 22} & $%
\begin{array}{c}
{\small 4.9548,} \\
{\small 2.9802}%
\end{array}%
$ &  &  \\ \hline
\multicolumn{6}{|c|}{{\small Notice: Runtime is 4.3411s. The associated
figure is Fig.\ref{figInstance2}}} \\ \hline
\end{tabular}

\subsection{Appendix III.}

$f_{3}(x_{1},x_{2})=x_{1}^{2}+x_{2}^{2}-\cos (18x_{1})-\cos (18x_{2}),$ $\ \
\ x_{1},x_{2}\in \lbrack -1,1]$

The interval $[-1,1]\otimes \lbrack -1,1]$ is divided into $n=40\times 40$
intervals initially. And at next iteration steps, search domain is divided
into $n1=20\times 20$ small intervals.

\bigskip

\begin{tabular}{|c|c|c|c|c|c|}
\hline
\multicolumn{6}{|l|}{\ \ \ \ \ \ \ \ \ \ \textbf{Table 5. The Extreme Points
of Function }$f_{3}(x_{1},x_{2})$} \\ \hline
\multicolumn{2}{|c|}{} &  &  &  &  \\ \hline
{\small Extreme} & {\small Calculated Value} & {\small Extreme} & {\small %
Calculated Value} & {\small Extreme} & {\small Calculated Value} \\ \hline
{\small Points} & $\left( x_{1},x_{2},f(x_{1,}x_{2})\right) $ & {\small %
Points} & $\left( x_{1},x_{2},f(x_{1,}x_{2})\right) $ & {\small Points} & $%
\left( x_{1},x_{2},f(x_{1,}x_{2})\right) $ \\ \hline
{\small 1} & $%
\begin{array}{c}
{\small -0.8781,-0.8781,} \\
{\small 3.5326}%
\end{array}%
$ & {\small 13} & $%
\begin{array}{c}
{\small -0.1810,-0.8725,} \\
{\small 2.7873}%
\end{array}%
$ & {\small 25} & $%
\begin{array}{c}
{\small 0.5175,-0.8700,} \\
{\small 3.0176}%
\end{array}%
$ \\ \hline
{\small 2} & $%
\begin{array}{c}
{\small -0.8737,-0.5310,} \\
{\small 3.0362}%
\end{array}%
$ & {\small 14} & $%
\begin{array}{c}
{\small -0.1773,-0.5252,} \\
{\small 2.3056}%
\end{array}%
$ & {\small 26} & $%
\begin{array}{c}
{\small 0.5200,-0.5200,} \\
{\small 2.5367}%
\end{array}%
$ \\ \hline
{\small 3} & $%
\begin{array}{c}
{\small -0.8725,-0.1810,} \\
{\small 2.7873}%
\end{array}%
$ & {\small 15} & $%
\begin{array}{c}
{\small -0.1756,-0.1756,} \\
{\small 2.0613}%
\end{array}%
$ & {\small 27} & $%
\begin{array}{c}
{\small 0.5215,-0.1700,} \\
{\small 2.2969}%
\end{array}%
$ \\ \hline
{\small 4} & $%
\begin{array}{c}
{\small -0.8700,0.1675,} \\
{\small 2.7759}%
\end{array}%
$ & {\small 16} & $%
\begin{array}{c}
{\small -0.1715,0.1715,} \\
{\small 2.0559}%
\end{array}%
$ & {\small 28} & $%
\begin{array}{c}
{\small 0.5252,0.1773,} \\
{\small 2.3056}%
\end{array}%
$ \\ \hline
{\small 5} & $%
\begin{array}{c}
{\small -0.8700,0.5175,} \\
{\small 3.0176}%
\end{array}%
$ & {\small 17} & $%
\begin{array}{c}
{\small -0.1700,0.5215,} \\
{\small 2.2969}%
\end{array}%
$ & {\small 29} & $%
\begin{array}{c}
{\small 0.5268,0.5268,} \\
{\small 2.5517}%
\end{array}%
$ \\ \hline
{\small 6} & $%
\begin{array}{c}
{\small -0.8675,0.8675,} \\
{\small 3.4966}%
\end{array}%
$ & {\small 18} & $%
\begin{array}{c}
{\small -0.1675,0.8700,} \\
{\small 2.7759}%
\end{array}%
$ & {\small 30} & $%
\begin{array}{c}
{\small 0.5310,0.8737,} \\
{\small 3.0362}%
\end{array}%
$ \\ \hline
{\small 7} & $%
\begin{array}{c}
{\small -0.5310,-0.8737,} \\
{\small 3.0362}%
\end{array}%
$ & {\small 19} & $%
\begin{array}{c}
{\small 0.1675,-0.8700,} \\
{\small 2.7759}%
\end{array}%
$ & {\small 31} & $%
\begin{array}{c}
{\small 0.8675,-0.8675,} \\
{\small 3.4966}%
\end{array}%
$ \\ \hline
{\small 8} & $%
\begin{array}{c}
{\small -0.5268,-0.5268,} \\
{\small 2.5517}%
\end{array}%
$ & {\small 20} & $%
\begin{array}{c}
{\small 0.1700,-0.5215,} \\
{\small 2.2969}%
\end{array}%
$ & {\small 32} & $%
\begin{array}{c}
{\small 0.8700,-0.5175,} \\
{\small 3.0176}%
\end{array}%
$ \\ \hline
{\small 9} & $%
\begin{array}{c}
{\small -0.5252,-0.1773,} \\
{\small 2.3056}%
\end{array}%
$ & {\small 21} & $%
\begin{array}{c}
{\small 0.1715,-0.1715,} \\
{\small 2.0559}%
\end{array}%
$ & {\small 33} & $%
\begin{array}{c}
{\small 0.8700,-0.1675,} \\
{\small 2.7759}%
\end{array}%
$ \\ \hline
{\small 10} & $%
\begin{array}{c}
{\small -0.5215,0.1700,} \\
{\small 2.2969}%
\end{array}%
$ & {\small 22} & $%
\begin{array}{c}
{\small 0.1756,0.1756,} \\
{\small 2.0613}%
\end{array}%
$ & {\small 34} & $%
\begin{array}{c}
{\small 0.8725,0.1810,} \\
{\small 2.7873}%
\end{array}%
$ \\ \hline
{\small 11} & $%
\begin{array}{c}
{\small -0.5200,0.5200,} \\
{\small 2.5367}%
\end{array}%
$ & {\small 23} & $%
\begin{array}{c}
{\small 0.1773,0.5252,} \\
{\small 2.3056}%
\end{array}%
$ & {\small 35} & $%
\begin{array}{c}
{\small 0.8737,0.5310,} \\
{\small 3.0362}%
\end{array}%
$ \\ \hline
{\small 12} & $%
\begin{array}{c}
{\small -0.5175,0.8700,} \\
{\small 3.0176}%
\end{array}%
$ & {\small 24} & $%
\begin{array}{c}
{\small 0.1810,0.8725,} \\
{\small 2.7873}%
\end{array}%
$ & {\small 36} & $%
\begin{array}{c}
{\small 0.8781,0.8781,} \\
{\small 3.5326}%
\end{array}%
$ \\ \hline
\multicolumn{6}{|c|}{{\small Notice: All points are shown at Fig.\ref%
{figInstance3}. Runtime is 3203.2968s. \ \ \ \ \ \ }} \\ \hline
\end{tabular}

\end{document}